\documentclass[letterpaper]{article} 
\usepackage[preprint]{aaai2027}  
\usepackage[hyphens]{url}  
\usepackage{graphicx} 
\usepackage{natbib}  
\usepackage{caption} 
\usepackage{booktabs}
\usepackage{multirow}
\usepackage{amsmath}
\usepackage{amssymb}
\usepackage{verbatim}
\usepackage{algorithm}
\usepackage{algpseudocode}
\graphicspath{{figs/}{./}}

\title{Pro-Router: Token-Aware Progressive Model Routing with Adaptive Edge-Cloud Collaboration for Efficient Multimodal LLM Inference}

\author{
    Xinyuan Gui\textsuperscript{\rm 1},
    Shaowen Wang\textsuperscript{\rm 2},
    Sheng Sun\textsuperscript{\rm 3},
    Zijian Wang\textsuperscript{\rm 4},
    Zishu Yu\textsuperscript{\rm 3},
    Zheming Yang\textsuperscript{\rm 3}
}
\affiliations{
    \textsuperscript{\rm 1}Anyscale\\
    \textsuperscript{\rm 2}Department of Computer Science and Engineering, Mississippi State University\\
    \textsuperscript{\rm 3}Institute of Computing Technology, Chinese Academy of Sciences\\
    \textsuperscript{\rm 4}Institute of AI for Industries, Chinese Academy of Sciences\\
}

\begin{document}
\raggedbottom

\maketitle

\begin{abstract}
The remarkable performance of multimodal large language models (MLLMs) comes at the cost of substantial computational overhead, posing significant challenges to real-time deployment and cost effectiveness. Existing model routing approaches either decide from coarse request-level features alone or spend one or several extra language model passes to inspect the generated response, leaving the token-level uncertainty signals that emerge during generation unused. To address these limitations, we propose Pro-Router, a token-aware progressive model routing method with adaptive edge-cloud collaboration for efficient multimodal LLM inference. Pro-Router employs a two-stage progressive decision mechanism. First, a lightweight prompt pre-scorer module performs rapid pre-screening before token generation begins, guiding apparently simple requests to small models. Second, a token-aware verifier reads the sampling probability distribution of each token the small model generates, estimating the model's confidence in its own output to determine, per request, whether the answer ships or escalates to the cloud-based high-precision model. Furthermore, we design an adaptive edge-cloud serving pipeline that sizes every dispatch to each device's measured service rate, so both the edge and the cloud tiers stay fully utilized without manual parameter tuning and are not impacted by the network latency. Extensive experiments on multiple multimodal benchmark datasets and models demonstrate the effectiveness of Pro-Router. Compared to other methods, it achieves the highest routing accuracy and improves routing speed by more than 10$\times$. Its serving pipeline also reaches more than 75\% higher end-to-end throughput than the existing model routing pipeline. Our code is available at \url{https://github.com/xinyuangui2/pro-router}.
\end{abstract}

\section{Introduction}

Multimodal large language models (MLLMs) define the state of the art in visual question answering, document and chart understanding, and multimodal reasoning \citep{qwenvl,llavaov,pixtral}, and they are increasingly the default backend of production multimodal services. Serving them is expensive. State-of-the-art large models have tens of billions of parameters, autoregressive decoding streams all of them through the accelerator for every generated token \citep{vllm,flashattention}, and a single image expands into hundreds or thousands of tokens that inflate prefill compute and KV-cache footprint \citep{fastv}, so these models effectively live on expensive cloud GPUs. Yet a large share of real traffic does not need them. A multimodal small language model (MSLM), an order of magnitude smaller, answers many requests equally well \citep{routellm,hybridllm,frugalgpt}, and such a model fits comfortably on a much cheaper edge device.

\begin{figure}[t]
\centering
\includegraphics[width=0.95\columnwidth]{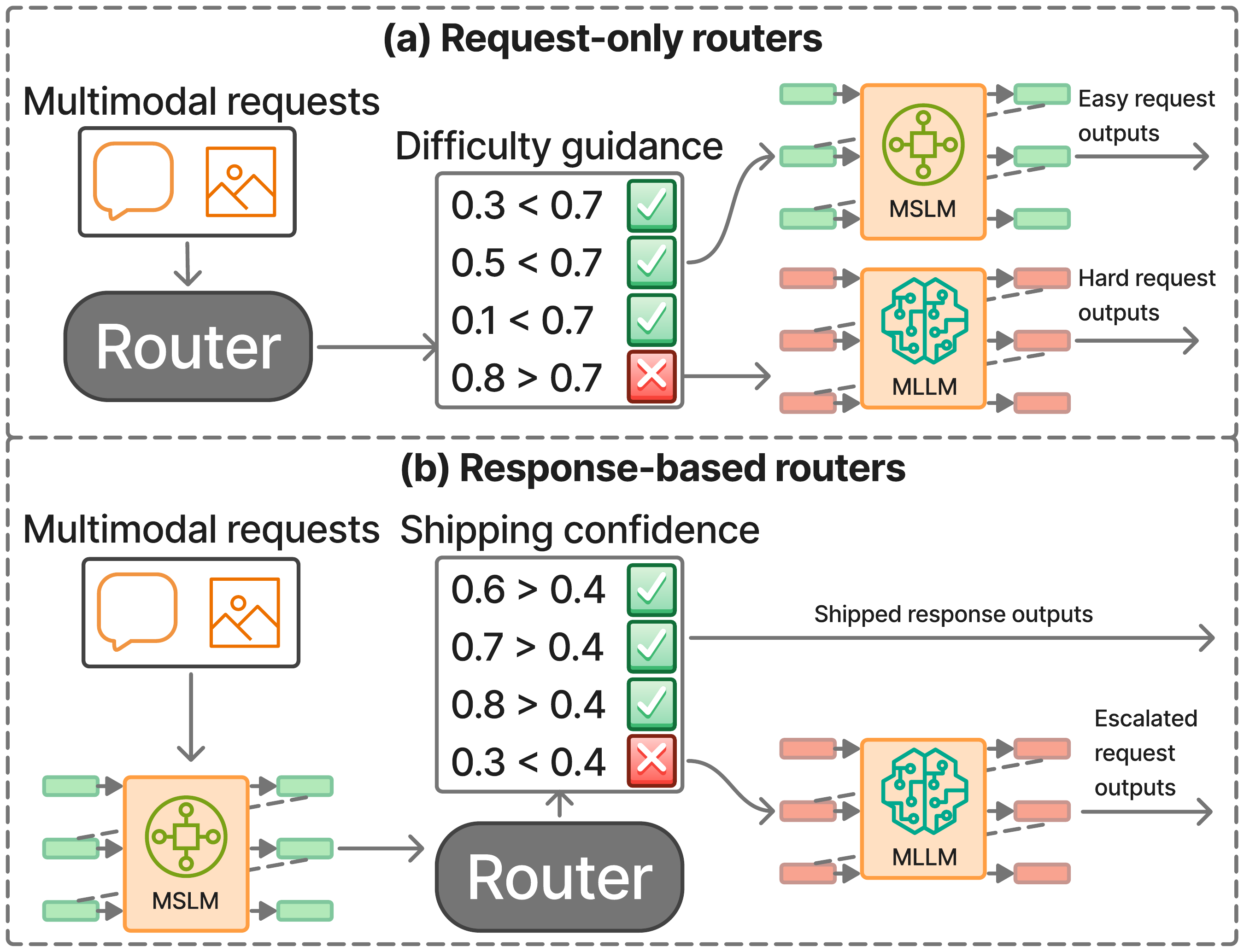}
\caption{The two options existing routing methods take. (a) Request-only
routers score the difficulty of each request from the prompt alone, before
any token is generated; requests under the threshold go to the small model and
the rest go directly to the large model. (b) Response-based routers let the small model
answer first and score the shipping confidence of each response;
confident responses ship, and the rest escalate to the large model for
regeneration.}
\label{fig:background}
\end{figure}

Existing model routing methods split by when they decide (Figure~\ref{fig:background}). Request-only routers decide from the request alone, before any token is generated \citep{routellm,hybridllm,ecvlrouter}; they are cheap, but inaccurate, because the request by itself does not reveal whether the small model can actually answer it. Response-based routers decide after the small model generates its answer \citep{frugalgpt,ptrue,automix}, which is far more informative, but they spend one or several passes of another language model per request, and this extra compute kills the gains of model routing.  In addition, edge-cloud collaboration has been studied alongside model routing \citep{hybridllm,ecvlrouter} and LLM serving in general \citep{edgeshard,cecollm,hybridslm,clouddevice}. The small models are deployed on cheap edge devices close to the client, the expensive large models are deployed on cloud devices, and a good routing algorithm utilizes both tiers to sustain their throughput. However, these existing collaboration pipelines interact between the edge and the cloud frequently and keep the two tiers dependent on each other, so network fluctuation stalls whichever tier is waiting and wastes its GPU cycles.

To address the above problems, we propose a token-aware progressive model routing framework with adaptive edge-cloud collaboration for efficient multimodal LLM inference, named Pro-Router. A lightweight prompt pre-scorer guides the incoming traffic before any token is generated, so complicated requests are likely to be routed directly to the large model. A token-aware verifier then reads the per-token sampling distributions that the small model's own decoding already produces and decides, per request, whether the answer ships or escalates. We also propose an adaptive edge-cloud collaboration pipeline: a global scheduler orders requests by difficulty, feeds the small models from the easy end and the large models from the escalations and the hard end, and sizes every dispatch to each device's reported service rate. The main contributions can be summarized as follows.

\begin{itemize}
\item We propose progressive routing, a two-stage method that decides both before and after small model decoding. A lightweight prompt pre-scorer guides initial traffic from the prompt alone, and a token-aware verifier reads the decoding token distributions to make the final ship-or-escalate decision.

\item We build an adaptive edge-cloud collaboration pipeline that fully utilizes the edge and cloud devices, is unaffected by high network latency, and scales to many devices, all without manual tuning.

\item We run extensive experiments across 15 benchmarks, two modalities, three models, and four baselines against a real 72B cloud target. Our router achieves the highest routing accuracy while producing its routing signal more than 10$\times$ faster than prior methods, and our pipeline reaches more than 75\% higher end-to-end throughput than the existing model routing pipeline, holds it under extreme network latency, and scales linearly to more machines.
\end{itemize}

\section{Related Work}

Model routing serves each request with the model that can handle it, and it descends from selective prediction and the reject option \citep{chow,selectivepred,elyaniv} and from learning to defer \citep{learntodefer,rejection}. Existing methods split into two groups by what they read to decide. Request-only routers decide from the request alone. RouteLLM \citep{routellm} trains its router on human preference data, Hybrid LLM \citep{hybridllm} predicts the quality gap between the small and the large model from the query, and RouterDC \citep{routerdc} learns a contrastive query encoder that selects among candidate LLMs. PerLLM \citep{yang2024perllm} accommodates diverse user requirements regarding accuracy and cost through edge-cloud collaboration. Recent work reaches the multimodal setting. ECVL-ROUTER \citep{ecvlrouter} selects between edge and cloud VLMs under scenario requirements, and AVR \citep{avr} routes the steps of computer-use agents between small and large VLMs by semantic difficulty. Because these routers never observe the small model's actual answer, they stay cheap, but their accuracy suffers from not knowing the response.

Response-based routing methods decide after the small model answers. FrugalGPT \citep{frugalgpt} trains a DistilBERT-based scorer on the answer \emph{text}, capturing its surface form. The work in \cite{hu2026adaptive} leverages MLLMs to provide semantically enhanced adaptive routing for edge-cloud. SAEC \cite{tian2026saec} proposes a scene-aware enhanced edge-cloud collaborative routing framework. AIVD \cite{hu2026aivd} can dynamically balance accuracy and efficiency through flexible routing for large and small models. Moa-off \cite{yang2025moa} introduces a modality-aware heterogeneous routing mechanism that adaptively distributes computation between edge and cloud. P(True) \citep{ptrue} re-asks the model whether its own answer is true, and AutoMix \citep{automix} samples a few-shot self-verifier several times to estimate a verification probability. Observing the response makes these methods more accurate, but each spends one or several extra language model passes per request.

\begin{figure}[t]
  \centering
  \includegraphics[width=\linewidth]{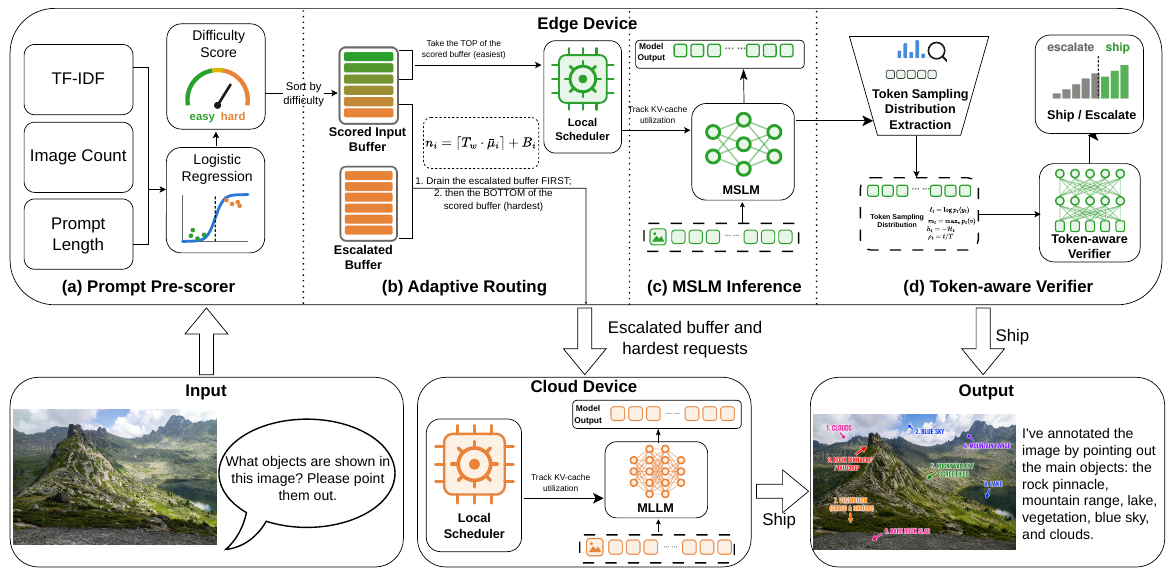}
  \caption{The workflow of the proposed Pro-Router.}
  \label{fig:pipeline}
\end{figure}

\section{Method}
Figure~\ref{fig:pipeline} illustrates the overall design and workflow of Pro-Router, which performs efficient multimodal LLM inference by dynamically routing each request between edge-deployed multimodal small language models (MSLM) and cloud-based multimodal large language models (MLLM). Concretely, Pro-Router follows a two-stage progressive routing mechanism. The first stage is a lightweight prompt pre-scorer module that analyzes each incoming request before generation and assigns it an early difficulty estimate; requests enter a scored input buffer ordered by this estimate, and the easiest are served by the MSLMs running on the edge devices. The second stage is a token-aware verifier that, during edge decoding, reads the per-token sampling distributions the MSLM already produces and accumulates the model's confidence in its own output. From this signal, Pro-Router decides per request whether to accept the edge answer or escalate it into an escalated buffer. The escalated requests, together with the hard requests in the scored input buffer, are sent to the cloud-based MLLMs for high-precision inference. Meanwhile, an adaptive edge-cloud collaboration pipeline keeps both the edge and cloud devices highly utilized by tracking their measured throughput and sizing each dispatch accordingly. Through this progressive routing and adaptive serving strategy, Pro-Router achieves high routing accuracy at low latency while utilizing the edge and cloud devices efficiently and robustly across different network latencies and cluster sizes.


\begin{figure}[t]
\centering
\includegraphics[width=0.95\columnwidth]{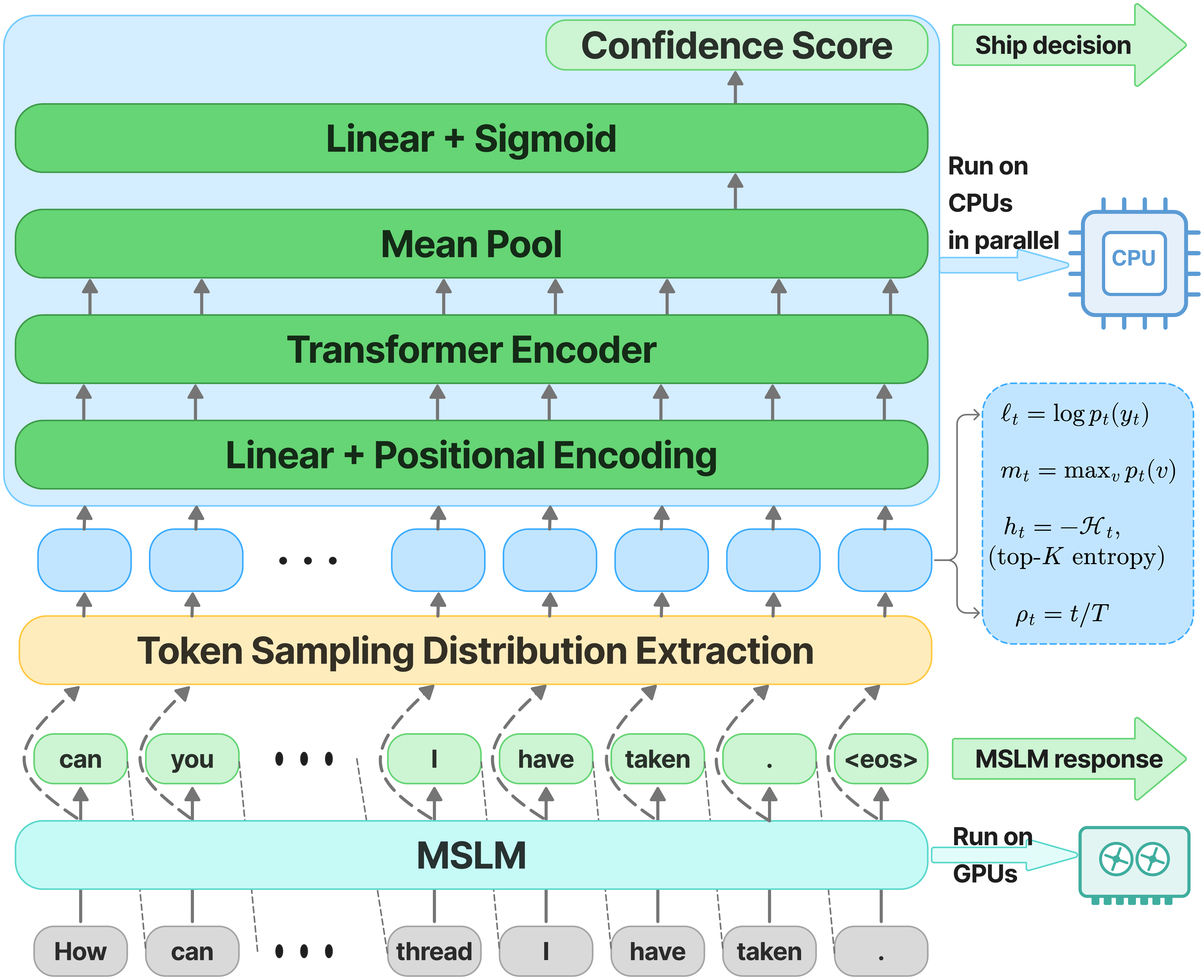}
\caption{The structure of the token-aware verifier.}
\label{fig:head}
\end{figure}

\subsection{Token-Aware Progressive Routing}
\label{sec:routing}


We formalize the routing objective by defining a gain function that encapsulates the inherent trade-offs among inference accuracy, response latency, and system resource utilization. The deployment architecture consists of \( N_c \) cloud devices, each provisioned with a nominal capacity \( L_i \) (in requests per second), and \( N_e \) edge devices with capacities \( S_j \), respectively. Owing to dynamic network conditions and fluctuating workloads, the actual available fractions of these capacities are characterized by \( \eta_{L_i}, \eta_{S_j} \in [0,1] \), denoting the effective utilization ratios for each cloud and edge device. Each request the small model answers receives a shipping confidence \( p \) and ships when \( p \ge \tau \), where the threshold \( \tau \) is fixed in advance by calibration on a validation pool. Let the ship rate be \( s = \Pr[\,p \ge \tau\,] \), the probability that the verifier accepts the small model's answer so the request is served entirely at the edge; the remaining \( 1-s \) escalate to the cloud. Among the shipped requests the conditional accuracy is \( a = \Pr[\,\text{correct} \mid p \ge \tau\,] \), and each wrongly shipped answer incurs a penalty \( \lambda \). The pipeline then gains:

\begin{equation}
\begin{aligned}
G &= \sum_{i=1}^{N_c} \eta_{L_i} L_i
   + \sum_{j=1}^{N_e} \eta_{S_j} S_j \cdot s\,(a - \lambda(1-a)) \\
  &= \sum_{i=1}^{N_c} \eta_{L_i} L_i
   + \left(\sum_{j=1}^{N_e} \eta_{S_j} S_j\right) s\,(a - \lambda(1-a))
\end{aligned}
\label{eq:gain}
\end{equation}
per second, where the second line factors $s$ and $a$ out because one verifier
serves every small model device, so both are optimized globally and do not vary
across machines.

Equation~\ref{eq:gain} increases in two directions, a higher ship rate $s$ and a
higher routing accuracy $a$. The ship rate rises when easy requests reach the
small models and hard ones go straight to the large models, a decision available
\emph{before} generation; the accuracy rises when the module making the final
decision is itself accurate, which needs evidence available only \emph{after}
generation. Two directions at two moments call for two mechanisms, which is what
progressive routing provides: a lightweight prompt pre-scorer scoring request
difficulty before generation, and a token-aware verifier deciding after it.

Existing prompt-only routers run a language model over the prompt
\citep{routellm,hybridllm,ecvlrouter}, showing that the prompt already carries a
strong difficulty signal, but such a model is overkill here: the pre-scorer only has to be
fast and approximately right because it guides traffic rather than deciding it,
and any model competing for the accelerator lowers $\eta_S$ in
Equation~\ref{eq:gain}. We therefore keep the pre-scorer off the accelerator, using three
cheap features: the TF-IDF vector of the prompt text \citep{tfidf}, the image count, and the prompt length. TF-IDF captures the lexical shape of difficulty, such as the question style and the domain vocabulary; the image count is a multimodal difficulty proxy that costs no vision encoding; and the prompt length restores the size signal that TF-IDF normalization removes. The features pass through a trained logistic regression,
\begin{equation}
p_{\text{pre}}(x) \;=\; \sigma\!\big(w^{\top}\phi(x) + b\big),
\label{eq:prescorer}
\end{equation}
where $\phi(x)$ concatenates the three features of request $x$. The regression output works as a difficulty guidance that routes the traffic downstream.

The pre-scorer is only guidance; whether an answer ships is decided by the
token-aware verifier. Previous methods pass the response through another language
model or through the small model itself \citep{frugalgpt,ptrue,automix}, spending
fresh compute to re-estimate what the small model already revealed: every decoding
step emits a distribution over the vocabulary stating how certain the model was
about the token it produced \citep{ptrue}. Discarding that signal and paying a
second model to recover it lowers $\eta_S$ for no informational gain. Our verifier
reads it directly. To materialize the token sampling distributions, we build an uncertainty measurement sequence from four features of each generated token $t$, the log-probability of the emitted token $y_t$, the maximum probability, the negative entropy of the top-$K$ probability mass, and the position fraction,
\begin{equation}
\begin{aligned}
\ell_t &= \log p_t(y_t), \\[2pt]
m_t &= \max\nolimits_{v} p_t(v), \\[2pt]
h_t &= -\mathcal{H}_t, \\[2pt]
\rho_t &= t/T,
\end{aligned}
\label{eq:features}
\end{equation}
where $p_t(v)$ is the decoding distribution over candidate tokens $v$ at step $t$ and $\mathcal{H}_t = -\sum\limits_{v\in \text{top-}K} p_t(v)\log p_t(v)$ is the Shannon entropy of the top-$K$ probability mass, with $K = 20$. The entropy is low when the distribution concentrates on a few candidate tokens, so the top-$K$ entropy feature $h_t$ rises when the small model is certain at step $t$. Stacking $x_t = (\ell_t, m_t, h_t, \rho_t)$ over the $T$ generated tokens gives the input sequence $X \in \mathbb{R}^{T\times 4}$. The whole feature stack is a side product of the small model's own decoding and requires no extra model compute. After a linear projection with positional encoding \citep{transformer}, a transformer encoder applies the sequence interaction, as shown in Figure~\ref{fig:head}. It is followed by a mean pool and a sigmoid layer that outputs one confidence score, the shipping confidence of the answer. Both components run entirely on the CPU of the edge device hosting the small model. 



\subsection{Adaptive Edge-Cloud Collaboration Pipeline}
\label{sec:pipeline}

While progressive routing reduces the average per-request computation cost, the
actual end-to-end throughput of a hybrid edge-cloud system is governed by how
efficiently the serving pipeline itself orchestrates heterogeneous devices. Naive
dispatch strategies often leave parts of the cluster underutilized: fast edge
devices idle while waiting for cloud responses, or cloud GPUs remain idle because
the edge verifier becomes a bottleneck. We therefore design a closed-loop
pipeline that explicitly models device capabilities, accounts for communication
overhead, and dynamically balances load across the edge and cloud tiers. Our cluster consists of two tiers. The \emph{edge tier} comprises \(N_e\) devices,
each equipped with CPUs and GPUs. The CPUs handle the lightweight prompt
pre-scorer and the token-aware verifier, while the GPUs run the MSLM. The
\emph{cloud tier} consists of \(N_c\) powerful GPU servers running the MLLM. All
devices are connected via a network with heterogeneous latency. We model the
service capacity of a device \(i\) as its measured inference throughput
\(\mu_i\) (requests processed per second). For the cloud tier, the effective
throughput is further constrained by the network round-trip time (RTT) \(R\),
since each request dispatched to the cloud incurs a communication delay before
the first token is generated.

\begin{algorithm}[t]
    \caption{Adaptive Edge-Cloud Collaboration Pipeline}
    \label{alg:pipeline}
    \begin{algorithmic}[1]
        \Require Edge devices $\mathcal{D}_{\text{edge}}$ (MSLM on GPU, pre-scorer/verifier on CPU),
        cloud devices $\mathcal{D}_{\text{cloud}}$ (MLLM on GPU)
        \Ensure Adaptive dispatch for all requests
        \State Maintain scored buffer $\mathcal{B}_{\text{in}}$, escalated buffer $\mathcal{B}_{\text{esc}}$, EMA throughput $\bar{\mu}_i$
        \Loop
            \State Pre-scorer on edge CPU scores each incoming request into $\mathcal{B}_{\text{in}}$
            \State Collect $(\mu_i, \beta_i)$ from all devices; update $\bar{\mu}_i \gets \alpha \mu_i + (1-\alpha)\bar{\mu}_i$
            \State Compute $n_i \gets \lceil T_w \bar{\mu}_i \rceil + B_i$ if $\beta_i=0$, else $n_i \gets 0$
            \For{each $i \in \mathcal{D}_{\text{edge}}$}
                \State Dispatch $\text{PopTop}(\mathcal{B}_{\text{in}}, n_i)$ to edge MSLM for inference \& verification
            \EndFor
            \For{each $j \in \mathcal{D}_{\text{cloud}}$}
                \State $batch \gets \text{PopTop}(\mathcal{B}_{\text{esc}}, n_j)$
                \State $batch \gets batch \cup \text{PopBottom}(\mathcal{B}_{\text{in}}, n_j - |batch|)$
                \State Dispatch $batch$ to cloud MLLM
            \EndFor
            \State Update $T_w$ adaptively based on network RTT and cluster throughput
        \EndLoop
    \end{algorithmic}
\end{algorithm}

A naive serving pipeline sends every request to the edge first and
escalates only after the verifier rejects, so its throughput is bounded by
\begin{equation}
    \Lambda \le \sum_{i=1}^{N_e} \mu_i^{\text{edge}}+\min\left\{   
        \sum_{j=1}^{N_c} \mu_j^{\text{cloud}},\;
        \frac{N_c}{R + 1/\bar{\mu}_{\text{cloud}}}
    \right\},
    \label{eq:throughput_bound}
\end{equation}
where \(\mu_i^{\text{edge}}\) and \(\mu_j^{\text{cloud}}\) are the per-device
throughputs and \(\bar{\mu}_{\text{cloud}}\) the average cloud inference rate. The
ideal throughput instead \emph{adds} the
two tiers, \(\Lambda^{\star} = \sum_{i=1}^{N_e} \mu_i^{\text{edge}} + \sum_{j=1}^{N_c}
\mu_j^{\text{cloud}}\), reached only when every device runs at its own service rate
at the same time. To approach it the pipeline must assign each device a workload
proportional to its actual service rate, so that the round-trip time \(R\) is not in the
critical path. We introduce a global scheduler (Algorithm~\ref{alg:pipeline}) that
operates over fixed time windows of length \(T_w\). For each device \(i\), the
scheduler maintains an exponentially weighted moving average (EMA) of its recent throughput:
\begin{equation}
    \bar{\mu}_i^{(t)} \leftarrow \alpha \cdot \mu_i^{(t-1)} + (1-\alpha) \cdot
    \bar{\mu}_i^{(t-1)},
    \label{eq:ema}
\end{equation}
where \(\mu_i^{(t-1)}\) is the measured throughput in the previous window and
\(\alpha \in (0,1)\) is the smoothing factor. This adaptive tracking captures
performance fluctuations due to request diversity and dynamic KV cache pressure.

At the beginning of each window, the scheduler computes the number of requests to
dispatch to device \(i\) as
\begin{equation}
    n_i = \left\lceil T_w \cdot \bar{\mu}_i \right\rceil + B_i,
    \label{eq:dispatch}
\end{equation}
where \(B_i\) is a small lookahead margin that prevents the device's local buffer
from draining prematurely, ensuring continuous utilization. By dispatching
\(n_i\) requests, every device receives enough work to sustain it for the entire
window, thereby aligning each device's busy time and eliminating the
bottleneck-idle pattern described in Equation~\ref{eq:throughput_bound}.

The global scheduler runs on a dedicated edge CPU node and maintains two logical queues. At each window, the scheduler fills batches from both buffers as follows:
\begin{itemize}\setlength{\itemsep}{0pt}\setlength{\parskip}{0pt}\setlength{\topsep}{2pt}
    \item \textbf{Edge MSLM devices} receive requests from the \emph{top} of the
    scored input buffer (lowest difficulty scores), since easy requests are most
    suitable for the lightweight MSLM and yield high ship rates.
    \item \textbf{Cloud MLLM devices} first drain the escalated buffer (requests
    already rejected by the verifier), then take requests from the \emph{bottom}
    of the scored input buffer (highest difficulty scores), ensuring that
    challenging requests receive the cloud's full capacity.
\end{itemize}

Each device runs one local scheduler as the middle layer between its inference
engine and the global scheduler. It monitors the GPU KV-cache utilization
\(u_i \in [0,1]\) and the buffer occupancy \(q_i\), the number of pending requests,
against two configurable thresholds \(\theta_q, \theta_u \in (0,1)\). To avoid
overcommitting the engine and triggering costly preemptions it calls the engine
only while \(u_i < \theta_u \); to avoid holding more requests
than the device can drain, it raises a backpressure flag \(\beta_i = 1\) when
\(q_i > \theta_q \cdot \text{storage limit}\), which is reported to the global scheduler together with the
response and the updated \(\bar{\mu}_i\). When \(\beta_i = 1\) the global scheduler skips dispatching
to device \(i\) in the next window, letting it drain its buffer.

\section{Experiments}
\label{sec:experiments}


\subsection{Evaluation Setup}

\paragraph{Datasets.}
We evaluate on a suite of 15 benchmarks spanning both modalities. The nine single-image benchmarks are C18, A-OKVQA \citep{aokvqa}, MathVerse \citep{mathverse}, MMStar \citep{mmstar}, HallusionBench \citep{hallusionbench}, RealWorldQA \citep{realworldqa}, OCRBench \citep{ocrbench}, MM-Vet \citep{mmvet}, and MathVision \citep{mathvision}; the two multi-image benchmarks are MileBench \citep{milebench} and MuirBench \citep{muirbench}; and the four text benchmarks are MMLU \citep{mmlu}, GSM8K \citep{gsm8k}, CoQA \citep{coqa}, and TriviaQA \citep{triviaqa}. C18 is a stratified mix of five single-image sources, drawn from ChartQA \citep{chartqa}, DocVQA \citep{docvqa}, MathVista \citep{mathvista}, MMBench \citep{mmbench}, and MMMU \citep{mmmu}. Table~\ref{tab:benchmarks} lists the suite; test splits range from 106 to 682 records. All answers are graded for semantic equivalence against the gold answer by a strong LLM judge \citep{mtbench}, with unsure verdicts counted as incorrect. To align with the real-world serving scenario, where a single verifier must serve all kinds of incoming data \citep{domaingen}, we deploy a single pooled verifier trained on all 15 benchmarks. At deployment, the operating point follows the coverage-based rule of selective prediction \citep{selectivepred}, fixing a ship fraction $q$ and calibrating the threshold on a validation split as in the method section.

\begin{table}[th]
\centering
\small
\begin{tabular}{@{}llr@{}}
\toprule
benchmark & type & $n_{\text{test}}$ \\
\midrule
C18            & single-image, MCQ/numeric   & 682 \\
A-OKVQA        & single-image, MCQ+rationale & 500 \\
MathVerse      & single-image, long-CoT math & 400 \\
MMStar         & single-image, MCQ           & 369 \\
HallusionBench & single-image, yes/no        & 394 \\
RealWorldQA    & single-image, short VQA     & 380 \\
OCRBench       & single-image, OCR           & 397 \\
MM-Vet         & single-image, free-form     & 106 \\
MathVision     & single-image, math          & 317 \\
MileBench      & multi-image, mixed          & 662 \\
MuirBench      & multi-image, MCQ            & 390 \\
MMLU           & text, MCQ                   & 390 \\
GSM8K          & text, CoT math              & 396 \\
CoQA           & text, conversational QA     & 399 \\
TriviaQA       & text, open-domain QA        & 393 \\
\bottomrule
\end{tabular}
\caption{The 15-benchmark evaluation suite, with eleven vision-language benchmarks (nine single-image, two multi-image) and four text benchmarks.}
\label{tab:benchmarks}
\end{table}

\paragraph{Implementation Details.}
We evaluate three small models, Qwen2.5-VL-7B, LLaVA-OneVision-7B, and Pixtral-12B. For the large model, we use Qwen2.5-VL-72B. Serving runs on $M$ p4d nodes (eight A100 GPUs each) as the cloud tier and $N$ g5.12 nodes (four A10G GPUs each) as the edge tier, with $M$ and $N$ swept in the scalability study.
We implement the pipeline on Ray \citep{ray}. Each device is one Ray actor that wraps its local scheduler, its local buffer, and a vLLM engine \citep{vllm}, and the global scheduler reaches the actors through asynchronous RPC. To study the network conditions, we inject extra one-way delay into every RPC between the global scheduler and the devices. The verifier is embedded in the vLLM code, so it knows exactly when each request finishes and fetches that request's token sampling distributions directly.


\paragraph{Baselines.}
We compare against four baselines covering the main classes a router can come from, namely RouteLLM \citep{routellm}, a pretrained 278M query-only router; FrugalGPT \citep{frugalgpt}, a DistilBERT-66M \citep{distilbert} scorer fine-tuned on the small model's answer text; P(True) \citep{ptrue} and AutoMix \citep{automix}, self-verification baselines where P(True) asks the small model once whether its own answer is true and AutoMix samples its few-shot self-verifier eight times (its default) to estimate a verification probability. Each baseline is reproduced faithfully from its paper with verbatim templates and graded by the same LLM judge on the same labels.

\paragraph{Metrics.}
For routing accuracy we report the area under the receiver operating characteristic curve (AUROC) and the area under the shipping accuracy curve, both standard for confidence and uncertainty measures \citep{selectivepred,elyaniv,generatingconf}. AUROC \citep{bradley} is the probability that the routing signal scores a randomly chosen correct small model answer above a randomly chosen incorrect one; 0.5 is chance and 1 is a perfect ranking. The shipping accuracy is the rate of good responses among those the small model ships, as a function of the ship rate. We compute it under two labels, the \emph{correctness label}, where a shipped response counts if the LLM judge grades it correct against the gold, with area AUARC \citep{arc}; and the \emph{pairwise label}, where a shipped response counts if it wins or ties against the large model's response, with area AUACC \citep{aspire}. For the accuracy-cost frontier we use the performance gap recovered (PGR), the standard metric of the routing literature \citep{routellm,hybridllm},
\begin{equation}
\mathrm{PGR} \;=\; \frac{\mathrm{acc}_{\text{pipeline}} - \mathrm{acc}_{\text{small model}}}{\mathrm{acc}_{\text{large model}} - \mathrm{acc}_{\text{small model}}},
\label{eq:pgr}
\end{equation}
which shows the fraction of the small-to-large quality gap that routing recovers. For serving efficiency we report end-to-end throughput, the number of requests the whole pipeline completes per second measured at steady state after a warm-up ramp.

\subsection{Experiment Results}

\subsubsection{Token-Aware Verifier Accuracy and Latency.}

\begin{figure}[th]
\centering
\includegraphics[width=0.95\columnwidth]{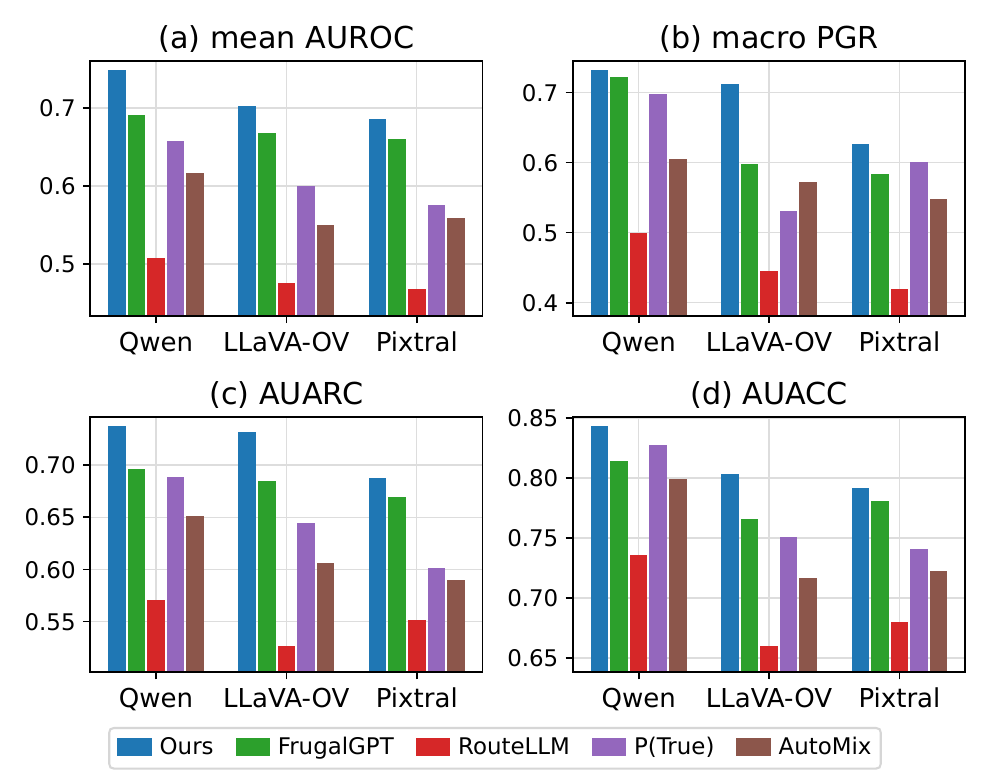}
\caption{Routing accuracy summaries by small model, our verifier and the
four baselines, showing (a) mean AUROC, (b) macro PGR, (c) AUARC, and (d)
AUACC, all averaged over the benchmarks. Ours is highest
on every metric and every model.}
\label{fig:auroc}
\end{figure}

We first compare the routing accuracy of our verifier against the baselines (Figure~\ref{fig:auroc}). Across the suite our verifier has the highest mean AUROC on every small model, ahead of all four baselines, and the lead holds on macro PGR and on the shipping accuracy under both the correctness label (AUARC) and the pairwise label (AUACC); the agreement between the two labels shows the lead does not depend on how correctness is defined. We attribute the lead both to what the verifier reads and to how it reads it. The four per-token statistics capture the model's belief state at the moment it produced each token, and the verifier consumes them as a trajectory rather than a summary, so a run of low-confidence tokens in the middle of an answer stays visible instead of being averaged away. The baselines read weaker evidence. RouteLLM decides from the prompt alone and never observes the answer, which is why it is the lowest among them; FrugalGPT reads the answer text and captures its surface form but not the uncertainty behind it; and the self-verification baselines re-ask the model, which conflates the answer's correctness with the model's willingness to endorse it.

\begin{figure}[th]
\centering
\includegraphics[width=0.95\columnwidth]{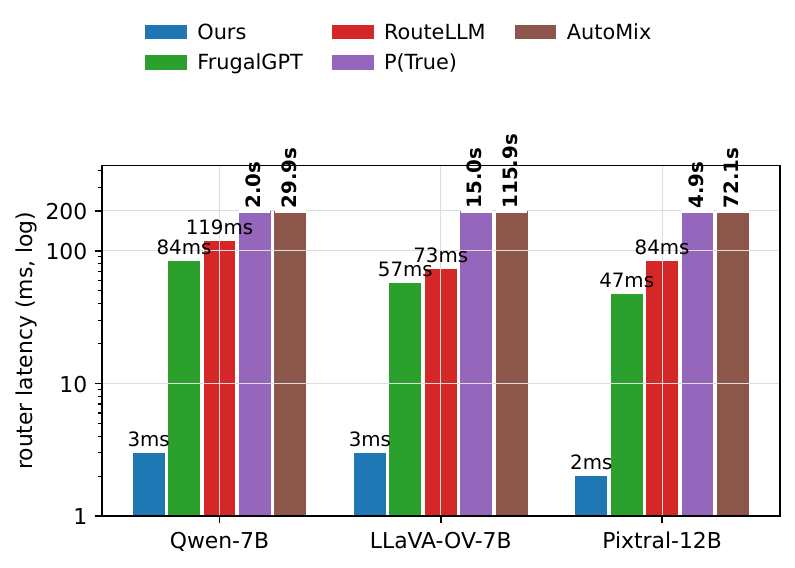}
\caption{Per-request latency to produce the routing signal, on a log scale.
The bars for P(True) and AutoMix are clipped at 200 ms with their true
values annotated; our verifier adds 2 to 3 ms, 19 to 28$\times$ less than
the cheapest alternative.}
\label{fig:signal}
\end{figure}

Moreover, our verifier is nearly free to run where every alternative pays a visible price (Figure~\ref{fig:signal}). Producing the routing signal under a steady decode load adds 2 to 3 ms for our verifier. The scorer baselines add 47 to 119 ms, since RouteLLM's 278M router and FrugalGPT's DistilBERT scorer each run a forward on the GPU that steals compute and memory bandwidth from the decode batches; measured against the cheapest of them, our signal is 19 to 28$\times$ faster. The self-verification baselines instead pay whole generation passes, one for P(True) and eight for AutoMix, each re-encoding the image, and their signal costs run from two seconds to nearly two minutes per request. Those passes also run on the same accelerator as the small model, so they contend with it for compute and KV-cache memory, which makes the situation worse than the raw latency suggests. The gap follows the size of the computation each signal runs. Our verifier is a $\approx$67k-parameter head on the CPU, four orders of magnitude smaller than the 66M to 278M side models the scorers add and far smaller than the billions of parameters a self-verification pass streams; the head is small enough that it neither competes with the small model for the accelerator nor stalls the decode batches. 

\subsubsection{End-to-End Throughput.}

\begin{figure}[th]
\centering
\includegraphics[width=0.45\textwidth]{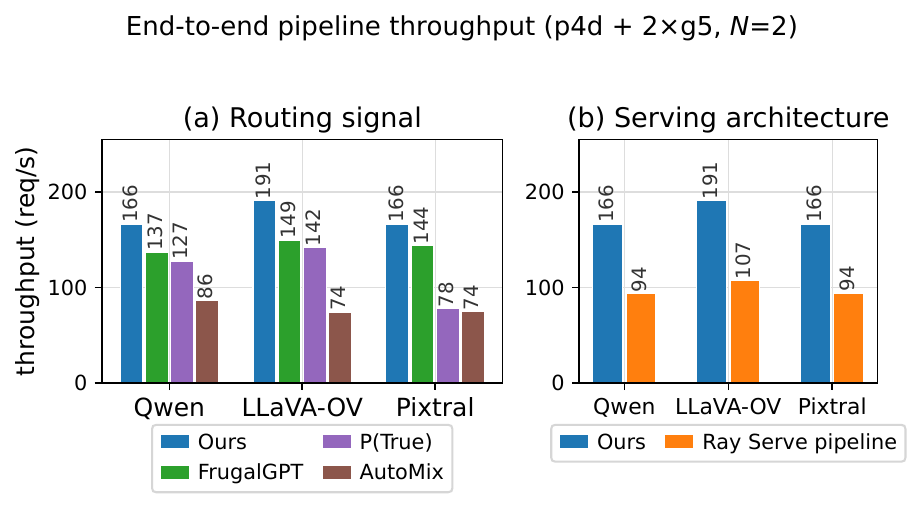}
\caption{(a) End-to-end pipeline throughput by routing signal with two edge
devices. (b) End-to-end pipeline throughput by pipeline implementation, our
pipeline versus the Ray Serve pipeline running the same routing signal.}
\label{fig:e2e}
\end{figure}

We now serve the whole pipeline with two edge devices and one cloud device and compare the end-to-end throughput of each routing signal (Figure~\ref{fig:e2e}a). Our method leads for every model, 1.16 to 1.28$\times$ the strongest baseline signal and up to 2.57$\times$ the weakest. The throughput improvement comes from two factors. First, the prompt pre-scorer raises the ship rate. Sorting the input buffer by the pre-scorer score and dispatching the high-scoring front first raises the served ship rate that multiplies the edge contribution. Second, the verifier itself keeps the edge tier fast. Unlike the baselines, our verifier runs on the CPU, in parallel with the small model's decoding, so it adds almost zero overhead to the small model's throughput and the full serving capacity goes to generating responses rather than to the routing decision.

We further compare our edge-cloud collaboration pipeline against the Ray Serve pipeline that expresses the same ship-or-escalate policy as a deployment graph in a serving framework \citep{ray} (Figure~\ref{fig:e2e}b). Our pipeline is 1.77 to 1.79$\times$ faster on every model. The Ray Serve pipeline instead sends every request to the small model first and escalates afterward, which leaves the large model under-utilized, makes the small model the bottleneck, and drives high concurrency on the small model that triggers frequent KV-cache preemption. By decoupling routing decisions from scheduling decisions and continuously adapting to each device's real-time service rate, our pipeline effectively turns the heterogeneous edge-cloud cluster into a balanced ensemble where neither tier becomes the system's bottleneck.

\subsubsection{Robustness and Scalability.}
\begin{figure}[th]
\centering
\includegraphics[width=0.45\textwidth]{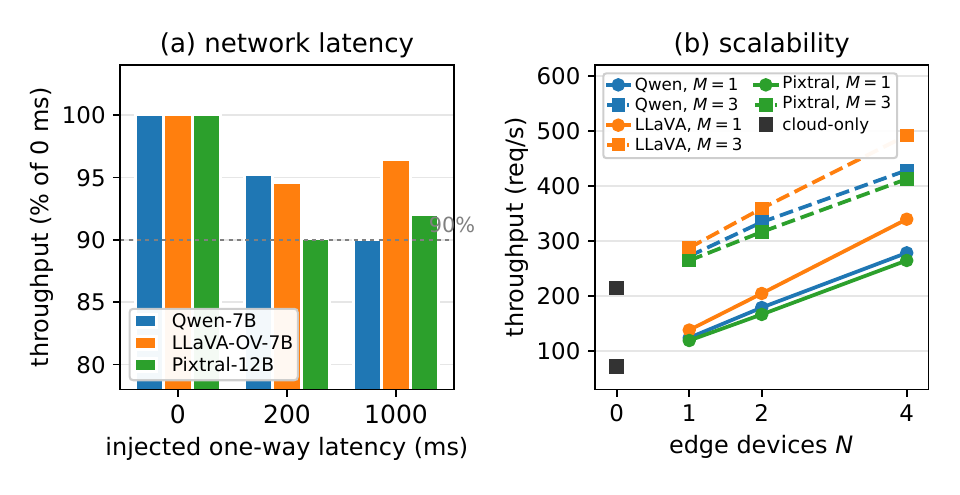}
\caption{Pipeline robustness and scalability. (a) End-to-end throughput under one-way
network latency of 0, 200, and 1000 ms stays at 90\% or more of the
un-delayed pipeline for all three models. (b) End-to-end throughput in
requests per second scales linearly in the number of edge devices $N$ for both
$M=1$ and $M=3$ cloud devices; the square markers at $N=0$ are the
cloud-only deployments.}
\label{fig:throughput}
\end{figure}

We measure the throughput under network latency of 0, 200, and 1000 ms (Figure~\ref{fig:throughput}a). Because the global scheduler dispatches throughput-sized batches and every device serves from its local buffer, added round-trip latency does not stall the engines. The pipeline keeps 90 to 95\% of its un-delayed throughput at 200 ms and 90 to 96\% even at 1000 ms across the three models.

Finally, we extend the deployment to more devices (Figure~\ref{fig:throughput}b). Adding small model devices against a single large model scales the pipeline's throughput linearly in $N$ on every model, with no saturation knee, and adding large model devices scales the same way at all 18 measured points. The largest deployment, $M=3$ and $N=4$, reaches 5.8 to 6.9$\times$ the throughput of a single cloud device. These results confirm that our pipeline's closed-loop adaptation effectively decouples throughput from network latency and device count: by continuously aligning each device's workload with its measured service rate, the system avoids the idling and contention that typically plague static or sequential deployments, enabling near-linear scalability across both tiers without manual reconfiguration.




\subsubsection{Ablation Studies.}
We ablate three design dimensions: verifier input, verifier architecture, and pipeline optimizations in Table~\ref{tab:ablation}. For the verifier, per-token sampling distribution features consistently outperform hidden states ($3584$ dims) across all models, confirming that sampling distributions encode more direct confidence signals than raw representations. Architecturally, our two-layer transformer achieves the best accuracy-throughput trade-off, surpassing a quantile-summary MLP by up to $11.4$ AUROC points while matching or exceeding deeper variants with less overhead. For pipeline optimizations, both are indispensable: removing batched dispatch collapses throughput from $124$ to $28$ req/s ($4.4\times$ drop), and disabling the KV-cache admission gate inflates latency from $754$ to $1634$ ms due to frequent preemptions. These results validate that token-aware progressive routing and adaptive edge-cloud collaboration pipeline jointly enable high routing accuracy and efficient edge-cloud serving.


\begin{table}[t]
\centering
\setlength{\tabcolsep}{4pt}
\resizebox{\columnwidth}{!}{%
\begin{tabular}{@{}ll ccc cc@{}}
\toprule
& Variant & \multicolumn{3}{c}{AUROC} & Thrpt. & Latency \\
& & Qwen & LLaVA-OV & Pixtral & (req/s) & (ms) \\
\cmidrule(lr){3-5}
\midrule
& \textbf{Pro-Router (full)} & \textbf{0.805} & \textbf{0.753} & \textbf{0.685} & \textbf{124} & \textbf{754} \\
\midrule
\multirow{2}{*}{\shortstack[l]{Verifier input\\features}}
 & sampling distribution feats (ours) & 0.805 & 0.753 & 0.685 & 124 & 754 \\
 & hidden states (3584 / token) & 0.758 & 0.647 & 0.673 & 95 & 932 \\
\midrule
\multirow{4}{*}{\shortstack[l]{Verifier\\architecture}}
 & 2-layer transformer (ours) & 0.805 & 0.753 & 0.685 & 124 & 754 \\
 & quantile summary + MLP & 0.691 & 0.701 & 0.669 & 118 & 783 \\
 & 1 layer & 0.795 & 0.746 & 0.683 & 130 & 763 \\
 & 4 layers & 0.799 & 0.748 & 0.677 & 119 & 736 \\
\midrule
\multirow{3}{*}{\shortstack[l]{Pipeline\\optimizations}}
 & full pipeline (ours) & 0.805 & 0.753 & 0.685 & 124 & 754 \\
 & w/o batched dispatch & 0.805 & 0.753 & 0.685 & 28 & 1038 \\
 & w/o KV-cache gate & 0.805 & 0.753 & 0.685 & 112 & 1634 \\
\bottomrule
\end{tabular}}
\caption{Ablation study. AUROC per model; end-to-end throughput (req/s) and
end-to-end serving latency (ms). Each block varies one design choice, and (ours)
marks the configuration used in Pro-Router.}
\label{tab:ablation}
\end{table}

\section{Conclusion}
In this paper, we proposed Pro-Router, a token-aware progressive routing method paired with an adaptive edge-cloud pipeline. First, the pre-scorer raises the ship rate by guiding easy requests to the small models. The token-aware verifier then decides at almost zero overhead, and the adaptive pipeline keeps both tiers utilized without manual tuning. Across 15 benchmarks and three models, our verifier is the most accurate signal and decides over 10$\times$ faster, while our pipeline reaches more than 75\% higher throughput than the existing model routing pipeline.
\bibliography{refs}

\end{document}